\documentclass{article} 
\usepackage{iclr2017_conference,times}
\usepackage{hyperref}
\usepackage{url}

\usepackage{bm}
\usepackage{amsmath}
\usepackage{amssymb}
\usepackage{subcaption}
\usepackage[pdftex]{graphicx}
\usepackage{xcolor,colortbl}
\newcommand{\sys}{\mbox{Bi-Directional Attention Flow}} 
\newcommand{\sysshort}{\mbox{\sc BiDAF}}

\newcolumntype{L}[1]{>{\raggedright\let\newline\\\arraybackslash\hspace{0pt}}m{#1}}
\newcolumntype{C}[1]{>{\centering\let\newline\\\arraybackslash\hspace{0pt}}m{#1}}
\newcolumntype{R}[1]{>{\raggedleft\let\newline\\\arraybackslash\hspace{0pt}}m{#1}}

\title{Bi-Directional Attention Flow \\
for Machine Comprehension}

\author{Minjoon Seo$^1$\thanks{The majority of the work was done while the author was interning at the Allen Institute for AI.}\qquad Aniruddha Kembhavi$^2$\qquad Ali Farhadi$^{1,2}$\qquad Hananneh Hajishirzi$^1$ \\
University of Washington$^1$, Allen Institute for Artificial Intelligence$^2$\\
\texttt{\{minjoon,ali,hannaneh\}@cs.washington.edu}, \texttt{\{anik\}@allenai.org}\\
}
\iclrfinalcopy 

\begin{document}

\maketitle

\begin{abstract}
Machine comprehension (MC), answering a query about a given context paragraph, requires modeling complex interactions between the context and the query. Recently, attention mechanisms have been successfully extended to MC. Typically these methods use attention to focus on a small portion of the context and summarize it with a fixed-size vector, couple attentions temporally, and/or often form a uni-directional attention. 
In this paper we introduce the \sys\ (\sysshort) network, a multi-stage hierarchical process that represents the context at different levels of granularity and uses bi-directional attention flow mechanism to obtain a query-aware context representation without early summarization. 
Our experimental evaluations show that our  model achieves the state-of-the-art results in Stanford Question Answering Dataset (SQuAD) and CNN/DailyMail cloze test. 

\end{abstract}

\section{Introduction}\label{sec:intro}
The tasks of machine comprehension (MC) and question answering (QA) have gained significant popularity over the past few years within the natural language processing and computer vision communities. Systems trained end-to-end now achieve promising results on a variety of tasks in the text and image domains. 
One of the key factors to the advancement has been the use of neural attention mechanism, which enables the system to focus on a targeted area within a context paragraph (for MC) or within an image (for Visual QA), that is most relevant to answer the question~\citep{memnn,antol2015vqa,xiong2016dynamic}. 
Attention mechanisms in previous works typically have one or more of the following characteristics. 
First, the computed attention weights are often used to extract the most relevant information from the context for answering the question by summarizing the context into a fixed-size vector.
Second, in the text domain, they are often temporally dynamic, whereby the attention weights at the current time step are a function of the attended vector at the previous time step. 
Third, they are usually uni-directional, wherein the query attends on the context paragraph or the image. 

In this paper, we introduce the \sys\  (\sysshort) network, a hierarchical multi-stage architecture for modeling the representations of the context paragraph at different levels of granularity (Figure~\ref{fig:model}). 
\sysshort\ includes character-level, word-level, and contextual embeddings, and uses bi-directional attention flow to obtain a query-aware context representation. 
Our attention mechanism offers following improvements to the previously popular attention paradigms. 
First, our attention layer is not used to summarize the context paragraph into a fixed-size vector. 
Instead, the attention is computed for every time step, and the attended vector at each time step, along with the representations from previous layers, is allowed to \emph{flow} through to the subsequent modeling layer.
This reduces the information loss caused by early summarization. 
Second, we use a memory-\emph{less} attention mechanism.
That is, while we iteratively compute attention through time as in~\cite{Bahdanau2014NeuralMT}, the attention at each time step is a function of only the query and the context paragraph at the current time step and does not directly depend on the attention at the previous time step.
We hypothesize that this simplification leads to the division of labor between the attention layer and the modeling layer.
It forces the attention layer to focus on learning the attention between the query and the  context, and enables the modeling layer to focus on learning the interaction within the query-aware context representation (the output of the attention layer).
It also allows the attention at each time step to be unaffected from incorrect attendances at previous time steps. 
Our experiments show that memory-less attention gives a clear advantage over dynamic attention. 
Third, we use attention mechanisms in both directions, query-to-context and context-to-query, which provide complimentary information to each other.

Our \sysshort\ model\footnote{Our code and interactive demo are available at: \url{allenai.github.io/bi-att-flow/}}  outperforms all previous approaches on the highly-competitive Stanford Question Answering Dataset (SQuAD) test set leaderboard at the time of submission.
With a modification to only the output layer, \sysshort\ achieves the state-of-the-art results on the CNN/DailyMail cloze test. 
We also provide an in-depth ablation study of our model on the SQuAD development set, visualize the intermediate feature spaces in our model, and analyse its performance as compared to a more traditional language model for machine comprehension~\citep{rajpurkar2016squad}.

\section{Model}\label{sec:model}

\begin{figure}[t]
\centering
\includegraphics[width=34pc]{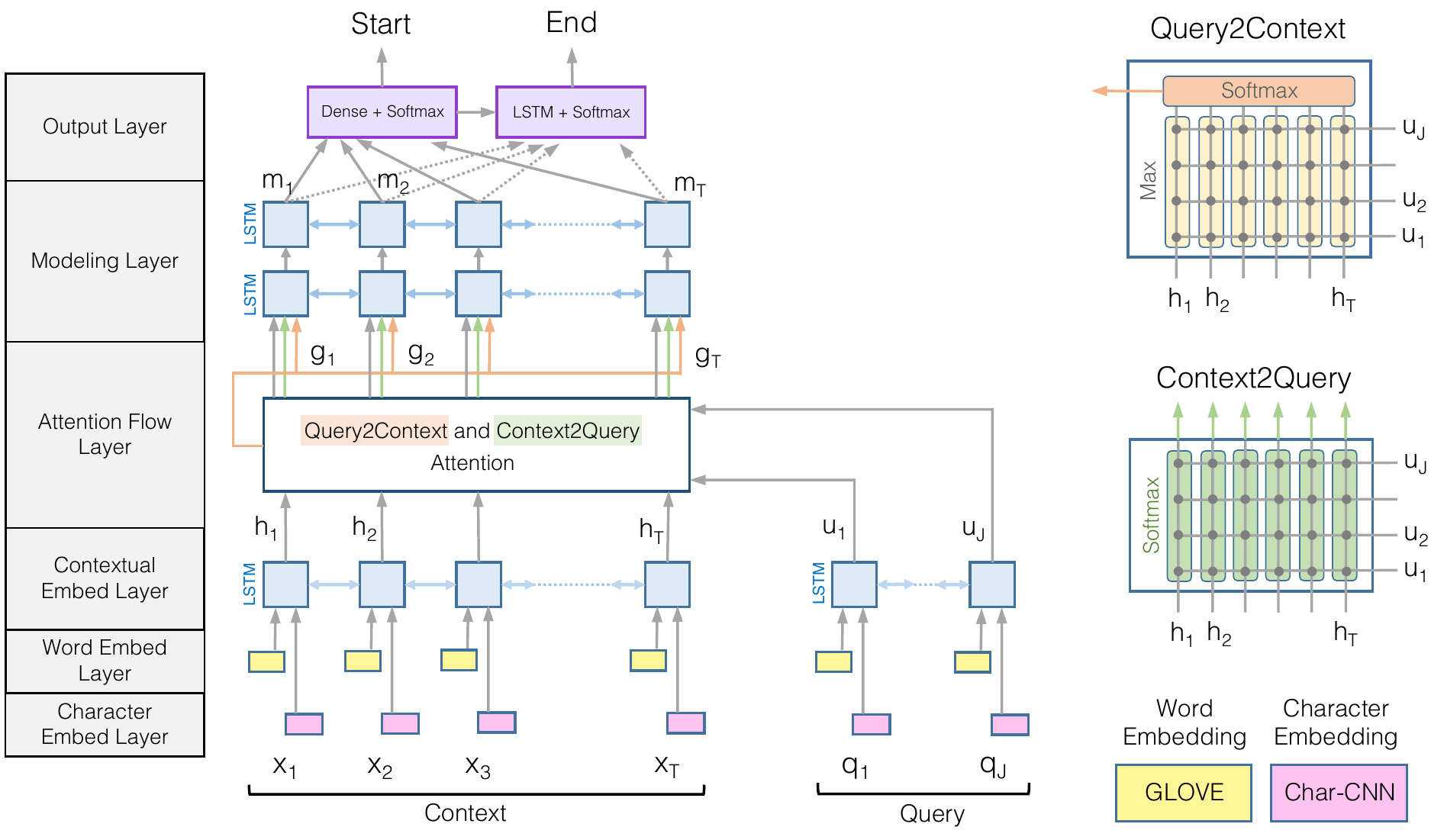}
\caption{\small BiDirectional Attention Flow Model\space\space \textit{(best viewed in color)}}
\label{fig:model}
\end{figure}
Our machine comprehension model is a hierarchical multi-stage process and consists of six layers (Figure~\ref{fig:model}): 
\begin{enumerate}
    \item \textbf{Character Embedding Layer} maps each word to a vector space using character-level CNNs. 
    \item \textbf{Word Embedding Layer} maps each word to a vector space using a pre-trained word embedding model. 
    \item \textbf{Contextual Embedding Layer} utilizes contextual cues from surrounding words to refine the embedding of the words. 
    These first three layers are applied to both the query and context.
    \item \textbf{Attention Flow Layer} 
    couples the query and context vectors and produces a set of query-aware feature vectors for each word in the context.
    \item \textbf{Modeling Layer} employs a Recurrent Neural Network to scan the context.
    \item \textbf{Output Layer} provides an answer to the query.
\end{enumerate}

\paragraph{1. Character Embedding Layer.}\label{subsec:char}
Character embedding layer is responsible for mapping each word to a high-dimensional vector space. Let $\{\bm{x}_1, \dots \bm{x}_T\}$ and $\{\bm{q}_1, \dots \bm{q}_J\}$ represent the words in the input context paragraph and query, respectively.
Following~\cite{char-cnn}, we obtain the character-level embedding of each word using Convolutional Neural Networks (CNN). Characters are embedded into vectors, which can be considered as 1D inputs to the CNN, and whose size is the input channel size of the CNN. The outputs of the CNN are max-pooled over the entire width to obtain a fixed-size vector for each word.  

\paragraph{2. Word Embedding Layer.}\label{subsec:emb}
Word embedding layer also maps each word to a high-dimensional vector space. We use pre-trained word vectors, GloVe~\citep{glove}, to obtain the fixed word embedding of each word.

The concatenation of the character and word embedding vectors is passed to a two-layer Highway Network~\citep{highway}. 
The outputs of the Highway Network are two sequences of $d$-dimensional vectors, or more conveniently, two matrices: ${\bf X} \in \mathbb{R}^{d \times T}$ for the context and ${\bf Q} \in \mathbb{R}^{d \times J}$ for the query.

\paragraph{3. Contextual Embedding Layer.}\label{subsec:pre}
We use a Long Short-Term Memory Network (LSTM)~\citep{lstm} on top of the embeddings provided by the previous layers to model the temporal interactions between words. 
We place an LSTM in both directions, and concatenate the outputs of the two LSTMs. Hence we obtain ${\bf H} \in \mathbb{R}^{2d \times T}$ from the context word vectors ${\bf X}$, and ${\bf U} \in \mathbb{R}^{2d \times J}$ from query word vectors ${\bf Q}$.
Note that each column vector of ${\bf H}$ and ${\bf U}$ is $2d$-dimensional because of the concatenation of the outputs of the forward and backward LSTMs, each with $d$-dimensional output.

It is worth noting that the first three layers of the model are computing features from the query and context at different levels of granularity, akin to the multi-stage feature computation of convolutional neural networks in the computer vision field.

\paragraph{4. Attention Flow Layer.}\label{subsec:att}
Attention flow layer is responsible for linking and fusing information from the context and the query words. 
Unlike previously popular attention mechanisms~\citep{memnn,hill2015goldilocks,iterative,reasonet}, the attention flow layer is not used to summarize the query and context into single feature vectors. 
Instead, the attention vector at each time step, along with the embeddings from previous layers, are allowed to flow through to the subsequent modeling layer. 
This reduces the information loss caused by early summarization.
 
The inputs to the layer are contextual vector representations of the context ${\bf H}$ and the query ${\bf U}$. 
The outputs of the layer are the query-aware vector representations of the context words, ${\bf G}$, along with the contextual embeddings from the previous layer.

In this layer, we compute attentions in two directions: from context to query as well as from query to context. 
Both of these attentions, which will be discussed below, are derived from a shared similarity matrix, ${\bf S} \in \mathbb{R}^{T \times J}$, between the contextual embeddings of the context (${\bf H}$) and the query (${\bf U}$), 
where ${\bf S}_{tj}$ indicates the similarity between $t$-th context word and $j$-th query word.
The similarity matrix is computed by
\begin{equation}\label{eqn:sim}
{\bf S}_{tj} = \alpha({\bf H}_{:t}, {\bf U}_{:j}) \in \mathbb{R}
\end{equation}
where $\alpha$ is a trainable scalar function that encodes the similarity between its two input vectors,
${\bf H}_{:t}$ is $t$-th column vector of ${\bf H}$, and
${\bf U}_{:j}$ is $j$-th column vector of ${\bf U}$,
We choose $\alpha({\bf h}, {\bf u}) = {\bf w}^\top_{({\bf S})} [{\bf h}; {\bf u}; {\bf h} \circ {\bf u}]$,
where ${\bf w}_{({\bf S})} \in \mathbb{R}^{6d}$ is a trainable weight vector, 
$\circ$ is elementwise multiplication,
$[;]$ is vector concatenation across row,
and implicit multiplication is matrix multiplication.
Now we use ${\bf S}$ to obtain the attentions and the attended vectors in both directions.

\textbf{\ \ \ Context-to-query Attention.} 
Context-to-query (C2Q) attention signifies which query words are most relevant to each context word.
Let ${\bf a}_t \in \mathbb{R}^J$ represent the attention weights on the query words by $t$-th context word, $\sum {\bf a}_{tj} = 1$ for all $t$. The attention weight is computed by ${\bf a}_t = \mathrm{softmax}({\bf S}_{t:}) \in \mathbb{R}^J$,
and subsequently each attended query vector is $\tilde{{\bf U}}_{:t} = \sum_j {\bf a}_{tj} {\bf U}_{:j}$.
Hence $\tilde{{\bf U}}$ is a $2d$-by-$T$ matrix containing the attended query vectors for the entire context.

\textbf{\ \ \ Query-to-context Attention.}
Query-to-context (Q2C) attention signifies which context words have the closest similarity to one of the query words and are hence critical for answering the query. We obtain the attention weights on the context words by ${\bf b} = \mathrm{softmax}(\max_{col} ({\bf S})) \in \mathbb{R}^T$, where the maximum function ($\max_{col}$) is performed across the column. Then the attended context vector is $\tilde{\bf h} = \sum_t {\bf b}_t {\bf H}_{:t} \in \mathbb{R}^{2d}$. This vector  indicates the weighted sum of the most important words in the context with respect to the query.
$\tilde{\bf h}$ is tiled $T$ times across the column, thus giving $\tilde{\bf H} \in \mathbb{R}^{2d \times T}$.

Finally, the contextual embeddings and the attention vectors are combined together to yield ${\bf G}$, where each column vector can be considered as the query-aware representation of each context word.
We define ${\bf G}$ by
\begin{equation}\label{eqn:flow}
{\bf G}_{:t} = {\bm \beta}({\bf H}_{:t}, \tilde{\bf U}_{:t}, \tilde{\bf H}_{:t}) \in \mathbb{R}^{d_{\bf G}}
\end{equation}
where ${\bf G}_{:t}$ is the $t$-th column vector (corresponding to $t$-th context word),
${\bm \beta}$ is a trainable vector function that fuses its (three) input vectors,
and $d_{\bf G}$ is the output dimension of the ${\bm \beta}$ function.
While the ${\bm \beta}$ function can be an arbitrary trainable neural network, such as multi-layer perceptron, 
a simple concatenation as following still shows good performance in our experiments: ${\bm \beta}({\bf h}, \tilde{\bf u}, \tilde{\bf h}) = [{\bf h}; \tilde{\bf u}; {\bf h} \circ \tilde{\bf u}; {\bf h} \circ \tilde{\bf h}] \in \mathbb{R}^{8d \times T}$ (i.e., $d_{\bf G} = 8d$).

\paragraph{5. Modeling Layer.}\label{subsec:main}
The input to the modeling layer is  ${\bf G}$, which encodes the query-aware representations of context words.
The output of the modeling layer captures the interaction among the context words conditioned on the query.
This is different from the contextual embedding layer, which captures the interaction among context words independent of the query. 
We use two layers of bi-directional LSTM, with the output size of $d$ for each direction. 
Hence we obtain a matrix ${\bf M}\in \mathbb{R}^{2d \times T}$, which is passed onto the output layer to predict the answer. 
Each column vector of ${\bf M}$ is expected to contain contextual information about the word with respect to the entire context paragraph and the query.

\paragraph{6. Output Layer.}\label{subsec:out}
The output layer is application-specific. 
The modular nature of \sysshort\ allows us to easily swap out the output layer based on the task, with the rest of the architecture remaining exactly the same. Here, we describe the output layer for the QA task. In section~\ref{subsec:cnn-details}, we use a slight modification of this output layer for cloze-style comprehension.  

The QA task requires the model to find a sub-phrase of the paragraph to answer the query. The phrase is derived by predicting the start and the end indices of the phrase in the paragraph. We obtain the probability distribution of the start index over the entire paragraph by
\begin{equation}
{\bf p}^{1} = \mathrm{softmax}({\bf w}_{({\bf p}^1)}^\top[{\bf G};{\bf M}]),
\end{equation}
where ${\bf w}_{({\bf p}^1)} \in \mathbb{R}^{10d}$ is a trainable weight vector.
For the end index of the answer phrase, we pass ${\bf M}$ to another bidirectional LSTM layer and obtain ${\bf M}^2 \in \mathbb{R}^{2d \times T}$.
Then we use ${\bf M}^2$ to obtain the probability distribution of the end index in a similar manner:
\begin{equation}
{\bf p}^{2} = \mathrm{softmax}({\bf w}_{({\bf p}^2)}^\top[{\bf G};{\bf M}^2])
\end{equation}

\textbf{Training.}
We define the training loss (to be minimized) as the sum of the negative log probabilities of the true start and end indices by the predicted distributions, averaged over all examples:
\begin{equation}
L(\theta) = -\frac{1}{N}\sum^N_i \log({\bf p}^1_{y^1_i}) + \log({\bf p}^2_{y^2_i})
\end{equation}
where $\theta$ is the set of all trainable weights in the model (the weights and biases of CNN filters and LSTM cells, ${\bf w}_{({\bf S})}$, ${\bf w}_{({\bf p}^1)}$ and ${\bf w}_{({\bf p}^2)}$),
$N$ is the number of examples in the dataset,
$y^1_i$ and $y^2_i$ are the true start and end indices of the $i$-th example, respectively,
and ${\bf p}_k$ indicates the $k$-th value of the vector ${\bf p}$.

\textbf{Test.}
The answer span $(k, l)$ where $k \leq l$ with the maximum value of ${\bf p}^1_k {\bf p}^2_l$ is chosen, which can be computed in linear time with dynamic programming.

\section{Related Work}\label{sec:related}
\paragraph{Machine comprehension.}
A significant contributor to the advancement of MC models has been the availability of large datasets. Early datasets such as MCTest~\citep{richardson2013mctest} were too small to train end-to-end neural models. 
Massive cloze test datasets (CNN/DailyMail by \citet{Hermann2015TeachingMT} and Childrens Book Test by \cite{hill2015goldilocks}), enabled the application of deep neural architectures to this task. 
More recently, \citet{rajpurkar2016squad} released the Stanford Question Answering (SQuAD) dataset with over 100,000 questions. We evaluate the performance of our comprehension system on both SQuAD and CNN/DailyMail datasets. 

Previous works in end-to-end machine comprehension use attention mechanisms in three distinct ways. 
The first group (largely inspired by~\cite{Bahdanau2014NeuralMT}) uses a dynamic attention mechanism, in which the attention weights are updated dynamically given the query and the context as well as the previous attention. 
\cite{Hermann2015TeachingMT} argue that the dynamic attention model performs better than using a single fixed query vector to attend on context words on CNN \& DailyMail datasets.
\cite{thorough} show that simply using bilinear term for computing the attention weights in the same model drastically improves the accuracy.
\cite{wang2016machine} reverse the direction of the attention (attending on query words as the context RNN progresses) for SQuAD. In contrast to these models, \sysshort\ uses a memory-less attention mechanism. 

The second group computes the attention weights once, which are then fed into an output layer for final prediction (e.g.,~\cite{kadlec2016text}).
Attention-over-attention model \citep{aoa} uses a 2D similarity matrix between the query and context words (similar to Equation~\ref{eqn:sim}) to compute the weighted average of query-to-context attention. In contrast to these models, \sysshort\ does not summarize the two modalities in the attention layer and instead lets the attention vectors flow into the modeling (RNN) layer. 

The third group (considered as variants of Memory Network~\citep{memnn}) repeats computing an attention vector between the query and the context through multiple layers, typically referred to as \emph{multi-hop} \citep{iterative,ga}. 
\cite{reasonet} combine Memory Networks with Reinforcement Learning in order to dynamically control the number of hops. One can also extend our \sysshort\ model to incorporate multiple hops.

\paragraph{Visual question answering.}
The task of question answering has also gained a lot of interest in the computer vision community. Early works on visual question answering (VQA) involved encoding the question using an RNN, encoding the image using a CNN and combining them to answer the question~\citep{antol2015vqa,Malinowski2015AskYN}. Attention mechanisms have also been successfully employed for the VQA task and can be broadly clustered based on the granularity of their attention and the approach to construct the attention matrix. At the coarse level of granularity, the question attends to different patches in the image~\citep{Zhu2015Visual7WGQ,xiong2016dynamic}. At a finer level, each question word attends to each image patch and the highest attention value for each spatial location~\citep{Xu2016AskAA} is adopted. A hybrid approach is to combine questions representations at multiple levels of granularity (unigrams, bigrams, trigrams)~\citep{yang2015stacked}. Several approaches to constructing the attention matrix have been used including element-wise product, element-wise sum, concatenation and Multimodal Compact Bilinear Pooling~\citep{fukui2016multimodal}.

\citet{lu2016hierarchical} have recently shown that in addition to attending from the question to image patches, attending from the image back to the question words provides an improvement on the VQA task. This finding in the visual domain is consistent with our finding in the language domain, where our bi-directional attention between the query and context provides improved results. 
Their model, however, uses the attention weights directly in the output layer and does not take advantage of the attention flow to the modeling layer.

\section{Question Answering Experiments}\label{sec:squad}
In this section, we evaluate our model on the task of question answering using the recently released SQuAD~\citep{rajpurkar2016squad}, which has gained a huge attention over a few months. In the next section, we evaluate our model on the task of cloze-style reading comprehension.

\paragraph{Dataset.}
SQuAD is a machine comprehension dataset on a large set of Wikipedia articles, with more than 100,000 questions. The answer to each question is always a span in the context. 
The model is given a credit if its answer matches one of the human written answers.  
Two metrics are used to evaluate models: Exact Match (EM) and a softer metric, F1 score, which measures the weighted average of the precision and recall rate at character level. 
The dataset consists of 90k/10k train/dev question-context tuples with a large hidden test set. 
It is one of the largest available MC datasets with human-written questions and serves as a great test bed for our model.

\paragraph{Model Details.}\label{subsec:squad-details}
The model architecture used for this task is depicted in Figure~\ref{fig:model}. Each paragraph and question are tokenized by a regular-expression-based word tokenizer (PTB Tokenizer) and fed into the model. We use 100 1D filters for CNN char embedding, each with a width of 5. 
The hidden state size ($d$) of the model is 100. 
The model has about 2.6 million parameters.
We use the AdaDelta~\citep{adadelta} optimizer, with a minibatch size of 60 and an initial learning rate of $0.5$, for 12 epochs. 
A dropout~\citep{dropout} rate of $0.2$ is used for the CNN, all LSTM layers, and the linear transformation before the softmax for the answers. 
During training, the moving averages of all weights of the model are maintained with the exponential decay rate of $0.999$. 
At test time, the moving averages instead of the raw weights are used.
The training process takes roughly 20 hours on a single Titan X GPU. We also train an ensemble model consisting of 12 training runs with the identical architecture and hyper-parameters. 
At test time, we choose the answer with the highest sum of confidence scores amongst the 12 runs for each question.

\paragraph{Results.} The results of our model and competing approaches on the hidden test  are summarized in Table~\ref{tab:squad-test}. \sysshort\ (ensemble) achieves an EM score of 73.3 and an F1 score of 81.1, outperforming all previous approaches.

\begin{table}[]
\begin{subfigure}[htbp]{0.6\textwidth}
    \centering
    \scalebox{0.9}{
    \begin{tabular}{lcccc}
        \hline
         & \multicolumn{2}{c}{Single Model} & \multicolumn{2}{c}{Ensemble} \\
         & EM & F1 & EM & F1\\
        \hline
        \hline
        Logistic Regression Baseline$^a$ & 40.4 & 51.0 & - & -\\
        Dynamic Chunk Reader$^{b}$ & 62.5 & 71.0 & - & -\\
        Fine-Grained Gating$^c$ & 62.5 & 73.3 & - & -\\
        Match-LSTM$^{d}$ & 64.7 & 73.7 & 67.9 & 77.0\\
        Multi-Perspective Matching$^e$ & 65.5 & 75.1 & 68.2 & 77.2 \\
        Dynamic Coattention Networks$^{f}$ & 66.2 & 75.9 & 71.6 & 80.4 \\
        R-Net$^{g}$ & \textbf{68.4} & \textbf{77.5} & 72.1 & 79.7\\
        \sysshort\ (Ours) & 68.0 & 77.3 & \textbf{73.3} & \textbf{81.1} \\
        \hline
        
    \end{tabular}
    }
    \caption{Results on the SQuAD test set}
    \label{tab:squad-test}
\end{subfigure}
\begin{subfigure}[htbp]{0.4\textwidth}
    \centering
    \scalebox{0.9}{
    \begin{tabular}{lcc}
        \hline
         & EM & F1\\
        \hline
        \hline
        No char embedding & 65.0 & 75.4\\
        No word embedding & 55.5 & 66.8\\
        No C2Q attention & 57.2 & 67.7 \\ 
        No Q2C attention & 63.6 & 73.7 \\
        Dynamic attention & 63.5 & 73.6\\
        \hline
        \sysshort\ (single) & 67.7 & 77.3 \\
        \sysshort\ (ensemble) & 72.6 & 80.7 \\
        \hline
    \end{tabular}
    }
    \caption{Ablations on the SQuAD dev set}
    \label{tab:squad-dev}
\end{subfigure}
\caption{(\ref{tab:squad-test}) The performance of our model \sysshort\ and competing approaches by \cite{rajpurkar2016squad}$^a$, \cite{chunk}$^b$, \cite{yang2016words}$^c$, \cite{wang2016machine}$^d$, IBM Watson$^e$ (unpublished), \cite{dcn}$^f$, and Microsoft Research Asia$^g$ (unpublished) on the SQuAD test set.
A concurrent work by \cite{lee2016learning} does not report the test scores. 
All results shown here reflect the SQuAD leaderboard (\url{stanford-qa.com}) as of 6 Dec 2016, 12pm PST. 
(\ref{tab:squad-dev}) The performance of our model and its ablations on the SQuAD dev set. Ablation results are presented only for single runs.}
\end{table}

\paragraph{Ablations.} Table~\ref{tab:squad-dev} shows the performance of our model and its ablations on the SQuAD dev set. Both char-level and word-level embeddings contribute towards the model's performance. We conjecture that word-level embedding is better at representing the semantics of each word as a whole, while char-level embedding can better handle out-of-vocab (OOV) or rare words. To evaluate bi-directional attention, we remove C2Q and Q2C attentions. For ablating C2Q attention, we replace the attended question vector $\tilde{\bf U}$ with the average of the output vectors of the question's contextual embedding layer (LSTM). C2Q attention proves to be critical with a drop of more than 10 points on both metrics. For ablating Q2C attention, the output of the attention layer, ${\bf G}$, does not include terms that have the attended Q2C vectors, $\tilde{\bf H}$. To evaluate the attention flow, we study a dynamic attention model, where the attention is dynamically computed within the modeling layer's LSTM, following previous work~\citep{Bahdanau2014NeuralMT,wang2016machine}. This is in contrast with our approach, where the attention is pre-computed before flowing to the modeling layer. Despite being a simpler attention mechanism, our proposed static attention outperforms the dynamically computed attention by more than 3 points. We conjecture that separating out the attention layer results in a richer set of features computed in the first 4 layers which are then incorporated by the modeling layer.
We also show the performance of \sysshort\ with several different definitions of $\alpha$ and ${\bm \beta}$ functions (Equation~\ref{eqn:sim} and~\ref{eqn:flow}) in Appendix~\ref{app:var}.

\paragraph{Visualizations.} We now provide a qualitative analysis of our model on the SQuAD dev set. First, we visualize the feature spaces after the word and contextual embedding layers. These two layers are responsible for aligning the embeddings between the query and context words which are the inputs to the subsequent attention layer. To visualize the embeddings, we choose a few frequent query words in the dev data and look at the context words that have the highest cosine similarity to the query words (Table~\ref{tab:viz_closest}). At the word embedding layer, query words such as \textit{When}, \textit{Where} and \textit{Who} are not well aligned to possible answers in the context, but this dramatically changes in the contextual embedding layer which has access to context from surrounding words and is just 1 layer below the attention layer. \textit{When} begins to match years, \textit{Where} matches locations, and \textit{Who} matches names. 

\begin{table}[]
    \scriptsize
    \centering
    \begin{tabular}{ lll }
        \hline
        \textbf{Layer} & \textbf{Query} & \textbf{Closest words in the Context using cosine similarity} \\
        \hline
        \hline
        Word & When & when, When, After, after, He, he, But, but, before, Before \\
        Contextual & When & When, when, 1945, 1991, 1971, 1967, 1990, 1972, 1965, 1953 \\
        \hline
        Word & Where & Where, where, It, IT, it, they, They, that, That, city \\
        Contextual & Where & where, Where, Rotterdam, area, Nearby, location, outside, Area, across, locations \\
        \hline
        Word & Who & Who, who, He, he, had, have, she, She, They, they \\
        Contextual & Who & who, whose, whom, Guiscard, person, John, Thomas, families, Elway, Louis \\
        \hline
        Word & city & City, city, town, Town, Capital, capital, district, cities, province, Downtown \\
        Contextual & city & city, City, Angeles, Paris, Prague, Chicago, Port, Pittsburgh, London, Manhattan \\
        \hline
        Word & January & July, December, June, October, January, September, February, April, November, March \\
        Contextual & January & January, March, December, August, December, July, July, July, March, December \\
        \hline
        Word & Seahawks & Seahawks, Broncos, 49ers, Ravens, Chargers, Steelers, quarterback, Vikings, Colts, NFL \\
        Contextual & Seahawks & Seahawks, Broncos, Panthers, Vikings, Packers, Ravens, Patriots, Falcons, Steelers, Chargers \\
        \hline
        Word & date & date, dates, until, Until, June, July, Year, year, December, deadline \\
        Contextual & date & date, dates, December, July, January, October, June, November, March, February \\
        \hline
    \end{tabular}
    \caption{\small Closest context words to a given query word, using a cosine similarity metric computed in the Word Embedding feature space and the Phrase Embedding feature space.}
    \label{tab:viz_closest}
\end{table}

\begin{figure}[t]
\centering
\includegraphics[width=34pc]{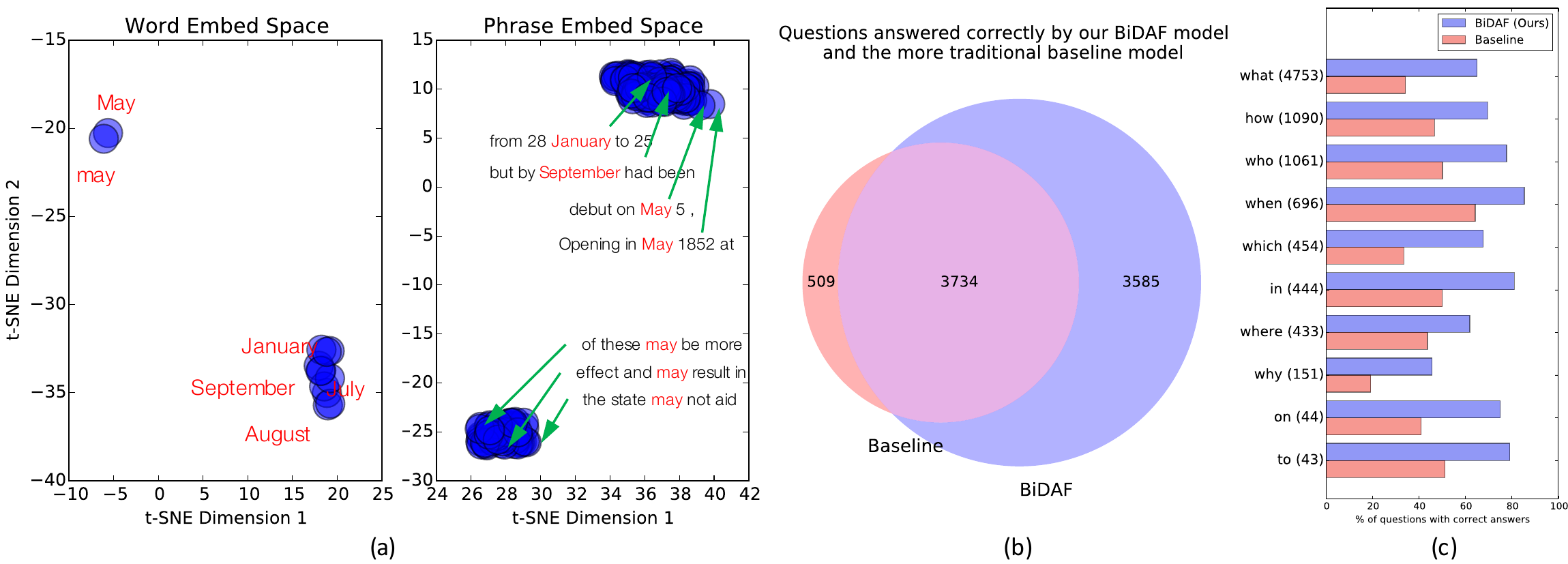}
\caption{\small (a) t-SNE visualizations of the \textit{months} names embedded in the two feature spaces. The contextual embedding layer is able to distinguish the two usages of the word \textit{May} using context from the surrounding text. (b) Venn diagram of the questions answered correctly by our model and the \textit{more traditional} baseline~\citep{rajpurkar2016squad}. (c) Correctly answered questions broken down by the 10 most frequent first words in the question.}
\label{fig:tsne}
\end{figure}

\begin{figure}[]
\centering
\includegraphics[width=30pc]{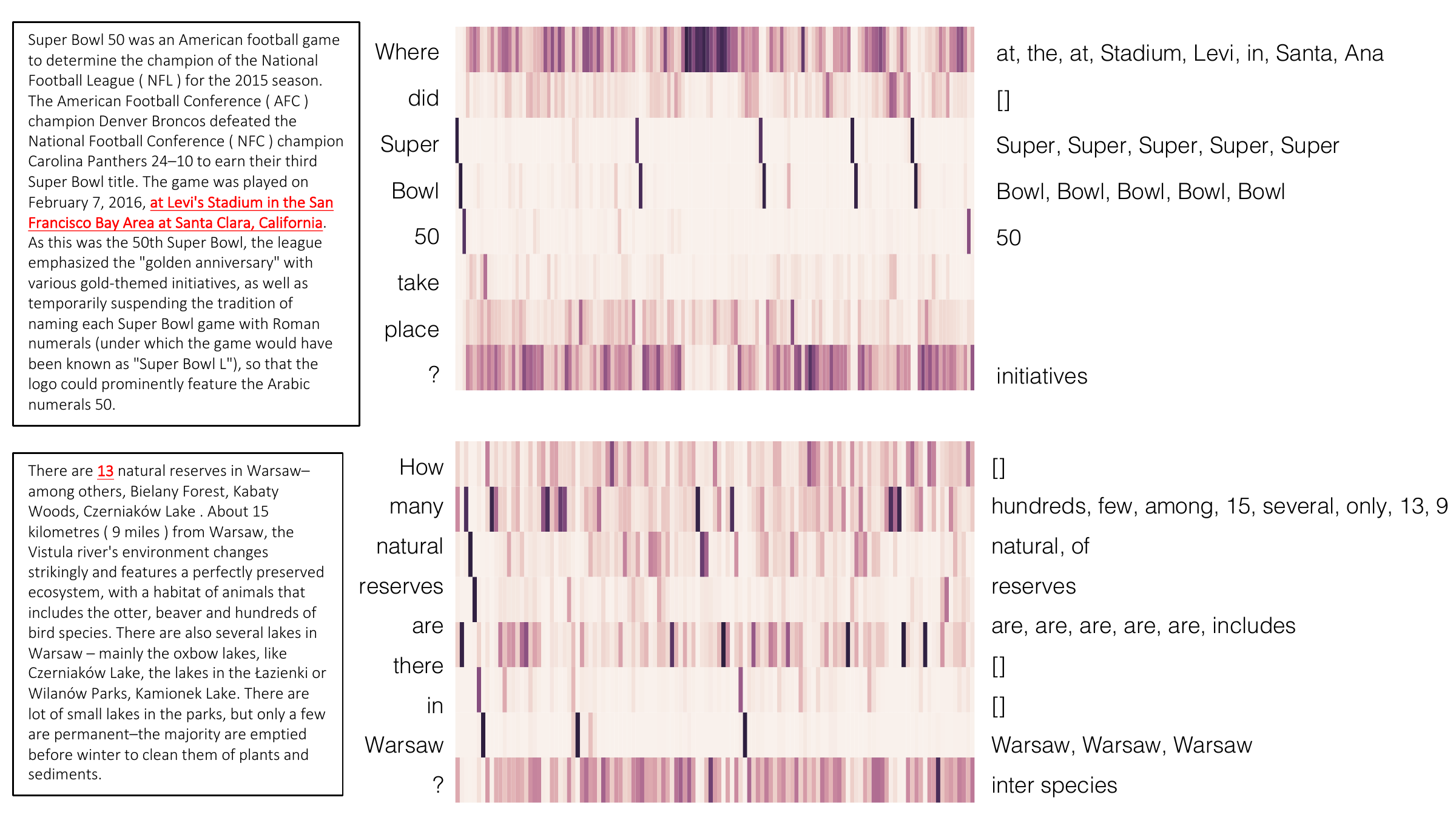}
\caption{\small Attention matrices for question-context tuples. The left palette shows the context paragraph (correct answer in red and underlined), the middle palette shows the attention matrix (each row is a question word, each column is a context word), and the right palette shows the top attention points for each question word, above a threshold.}
\label{fig:viz_attention}
\end{figure}

We also visualize these two feature spaces using t-SNE in Figure~\ref{fig:tsne}. t-SNE is performed on a large fraction of dev data but we only plot data points corresponding to the months of the year. 
An interesting pattern emerges in the Word space, where \textit{May} is separated from the rest of the months because \textit{May} has multiple meanings in the English language. 
The contextual embedding layer uses contextual cues from surrounding words and is able to separate the usages of the word \textit{May}. Finally we visualize the attention matrices for some question-context tuples in the dev data in Figure~\ref{fig:viz_attention}. In the first example, \textit{Where} matches locations and in the second example, \textit{many} matches quantities and numerical symbols. Also, entities in the question typically attend to the same entities in the context, thus providing a feature for the model to localize possible answers.

\paragraph{Discussions.} We analyse the performance of our our model with a traditional language-feature-based baseline~\citep{rajpurkar2016squad}. Figure~\ref{fig:tsne}b shows a Venn diagram of the dev set questions correctly answered by the models. Our model is able to answer more than 86\% of the questions correctly answered by the baseline. The 14\% that are incorrectly answered does not have a clear pattern. 
This suggests that neural architectures are able to exploit much of the information captured by the language features. 
We also break this comparison down by the first words in the questions (Figure~\ref{fig:tsne}c). Our model outperforms the traditional baseline comfortably in every category.

\paragraph{Error Analysis.} We randomly select 50 incorrect questions (based on EM) and categorize them into 6 classes.
50\% of errors are due to the imprecise boundaries of the answers,
28\% involve syntactic complications and ambiguities,
14\% are paraphrase problems,
4\% require external knowledge,
2\% need multiple sentences to answer,
and 2\% are due to mistakes during tokenization.
See Appendix~\ref{sec:error} for the examples of the error modes.

\section{Cloze Test Experiments}\label{sec:cnn}
We also evaluate our model on the task of cloze-style reading comprehension using the CNN and Daily Mail datasets~\citep{Hermann2015TeachingMT}.

\paragraph{Dataset.}
In a cloze test, the reader is asked to fill in words that have been removed from a passage, for measuring one's ability to comprehend text. \citet{Hermann2015TeachingMT} have recently compiled a massive Cloze-style comprehension dataset, consisting of 300k/4k/3k and 879k/65k/53k (train/dev/test) examples from CNN and DailyMail news articles, respectively. Each example has a news article and an incomplete sentence extracted from the human-written summary of the article. To distinguish this task from language modeling and force one to refer to the article to predict the correct missing word, the missing word is always a named entity, anonymized with a random ID. 
Also, the IDs must be shuffled constantly during test, which is also critical for full anonymization.

\paragraph{Model Details.}\label{subsec:cnn-details}
The model architecture used for this task is very similar to that for SQuAD (Section~\ref{sec:squad}) with only a few small changes to adapt it to the cloze test. 
Since each answer in the CNN/DailyMail datasets is always a single word (entity), we only need to predict the start index (${\bf p}^1$); the prediction for the end index (${\bf p}^2$) is omitted from the loss function. 
Also, we mask out all non-entity words in the final classification layer so that they are forced to be excluded from possible answers. Another important difference from SQuAD is that the answer entity might appear more than once in the context paragraph. 
To address this, we follow a similar strategy from~\cite{kadlec2016text}. 
During training, after we obtain ${\bf p}^1$, we sum all probability values of the entity instances in the context that correspond to the correct answer. 
Then the loss function is computed from the summed probability.
We use a minibatch size of 48 and train for 8 epochs, with early stop when the accuracy on validation data starts to drop. 
Inspired by the window-based method~\citep{hill2015goldilocks}, 
we split each article into short sentences where each sentence is a 19-word window around each entity (hence the same word might appear in multiple sentences).
The RNNs in \sysshort\ are not feed-forwarded or back-propagated across sentences, which speed up the training process by parallelization.
The entire training process takes roughly 60 hours on eight Titan X GPUs. The other hyper-parameters are identical to the model described in Section~\ref{subsec:squad-details}.

\paragraph{Results.}


The results of our single-run models and competing approaches on the CNN/DailyMail datasets are summarized in Table~\ref{tab:cnn}. 
$^*$ indicates ensemble methods. 
\sysshort\ outperforms previous single-run models on both datasets for both val and test data. On the DailyMail test, our single-run model even outperforms the best ensemble method.


\begin{table}[!htb]
    \centering
    \scalebox{0.9}{
    \begin{tabular}{lcccc}
        \hline
         & \multicolumn{2}{c}{CNN} & \multicolumn{2}{c}{DailyMail} \\
         & val & test & val & test\\
        \hline
        \hline
        Attentive Reader~\citep{Hermann2015TeachingMT} & 61.6 & 63.0 & 70.5 & 69.0 \\
        MemNN~\citep{hill2015goldilocks} & 63.4 & 6.8 & - & - \\
        AS Reader~\citep{kadlec2016text} & 68.6 & 69.5 & 75.0 & 73.9 \\
        DER Network~\citep{DER} & 71.3 & 72.9 & - & -\\
        Iterative Attention~\citep{iterative} & 72.6 & 73.3 & - & - \\
        EpiReader~\citep{epireader} & 73.4 & 74.0 & - & - \\
        Stanford AR~\citep{thorough} & 73.8 & 73.6 & 77.6 & 76.6 \\
        GAReader~\citep{ga} & 73.0 & 73.8 & 76.7 & 75.7 \\
        AoA Reader~\citep{aoa} & 73.1 & 74.4 & - & - \\
        ReasoNet~\citep{reasonet} & 72.9 & 74.7 & 77.6 & 76.6 \\
        \sysshort~(Ours) & \textbf{76.3} & \textbf{76.9} & \textbf{80.3} & \textbf{79.6} \\
        \hline
        MemNN$^*$~\citep{hill2015goldilocks} & 66.2 & 69.4 & - & - \\
        ASReader$^*$~\citep{kadlec2016text} & 73.9 & 75.4 & 78.7 & 77.7 \\
        Iterative Attention$^*$~\citep{iterative} & 74.5 & 75.7 & - & - \\
        GA Reader$^*$~\citep{ga} & 76.4 & 77.4 & 79.1 & 78.1\\
        Stanford AR$^*$~\citep{thorough} & 77.2 & 77.6 & 80.2 & 79.2 \\
        \hline
    \end{tabular}
    }
    \caption{\small Results on CNN/DailyMail datasets. We also include the results of previous ensemble methods (marked with $^*$) for completeness.}
    \label{tab:cnn}
\end{table}

\section{Conclusion}
In this paper, we introduce \sysshort, a multi-stage hierarchical process that represents the context at different levels of granularity and uses a bi-directional attention flow mechanism to achieve a query-aware context representation without early summarization. The experimental evaluations show that our  model achieves the state-of-the-art results in Stanford Question Answering Dataset (SQuAD) and CNN/DailyMail cloze test. The ablation analyses demonstrate the importance of each component in our model. The visualizations and discussions show that our model is learning a suitable representation for MC and is capable of answering complex questions by attending to correct locations in the given paragraph. Future work involves extending our approach to incorporate multiple hops of the attention layer.

\subsubsection*{Acknowledgments}
This research was supported by the NSF (IIS 1616112), NSF (III 1703166), Allen Institute for AI (66-9175), Allen Distinguished Investigator Award, Google Research Faculty Award, and Samsung GRO Award. We thank the anonymous reviewers for their helpful comments.

\newpage
\bibliography{00-main}
\bibliographystyle{iclr2017_conference}

\newpage
\appendix
\section{Error Analysis}\label{sec:error}
Table~\ref{tab:error} summarizes the modes of errors by \sysshort\  and shows examples for each category of error in SQuAD. 
\begin{table}[!htb]
    \centering
    \begin{tabular}{L{2cm} C{1.5cm} L{9.5cm}}
        Error type & Ratio (\%) & Example\\
        \hline
        \hline
        Imprecise answer boundaries & 50 & \parbox[t]{9cm}{\textbf{Context}: ``The Free Movement of Workers Regulation articles 1 to 7 set out the main provisions on equal treatment of workers.''\\
        \textbf{Question}: ``Which articles of the Free Movement of Workers Regulation set out the primary provisions on equal treatment of workers?''\\
        \textbf{Prediction}: ``1 to 7'',
        \textbf{Answer}: ``articles 1 to 7'' \\}\\
        \hline
        Syntactic complications and ambiguities & 28 & \parbox[t]{9cm}{\textbf{Context}: ``A piece of paper was later found on which Luther had written his last statement. ''\\
        \textbf{Question}: ``What was later discovered written by Luther?''\\
        \textbf{Prediction}: ``A piece of paper'',
        \textbf{Answer}: ``his last statement'' \\}\\
        \hline
        Paraphrase problems & 14 & \parbox[t]{9cm}{\textbf{Context}: ``Generally, education in Australia follows the three-tier model which includes primary education (primary schools), followed by secondary education (secondary schools/high schools) and tertiary education (universities and/or TAFE colleges).''\\
        \textbf{Question}: ``What is the first model of education, in the Australian system?''\\
        \textbf{Prediction}: ``three-tier'',
        \textbf{Answer}: ``primary education'' \\}\\
        \hline
        External knowledge & 4 & \parbox[t]{9cm}{\textbf{Context}: ``On June 4, 2014, the NFL announced that the practice of branding Super Bowl games with Roman numerals, a practice established at Super Bowl V, would be temporarily suspended, and that the game would be named using Arabic numerals as Super Bowl 50 as opposed to Super Bowl L.''\\
        \textbf{Question}: ``If Roman numerals were used in the naming of the 50th Super Bowl, which one would have been used?'\\
        \textbf{Prediction}: ``Super Bowl 50'',
        \textbf{Answer}: ``L''\\}\\
        \hline
        Multi-sentence & 2 & \parbox[t]{9cm}{\textbf{Context}: ``Over the next several years in addition to host to host interactive connections the network was enhanced to support terminal to host connections, host to host batch connections (remote job submission, remote printing, batch file transfer), interactive file transfer, gateways to the Tymnet and Telenet public data networks, X.25 host attachments, gateways to X.25 data networks, Ethernet attached hosts, and eventually TCP/IP and additional public universities in Michigan join the network. All of this set the stage for Merit's role in the NSFNET project starting in the mid-1980s.''\\
        \textbf{Question}: ``What set the stage for Merits role in NSFNET''\\
        \textbf{Prediction}: ``All of this set the stage for Merit 's role in the NSFNET project starting in the mid-1980s'',
        \textbf{Answer}: ``Ethernet attached hosts, and eventually TCP/IP and additional public universities in Michigan join the network''\\}\\
        \hline
        Incorrect preprocessing & 2 & \parbox[t]{9cm}{\textbf{Context}: ``English chemist John Mayow (1641-1679) refined this work by showing that fire requires only a part of air that he called spiritus nitroaereus or just nitroaereus.''\\
        \textbf{Question}: ``John Mayow died in what year?''\\
        \textbf{Prediction}: ``1641-1679'',
        \textbf{Answer}: ``1679''}\\
        
    \end{tabular}
    \caption{Error analysis on SQuAD. We randomly selected EM-incorrect answers and classified them into 6 different categories.
    Only relevant sentence(s) from the context shown for brevity. }
    \label{tab:error}
\end{table}

\section{Variations of Similarity and Fusion Functions}\label{app:var}
\begin{table}[htbp] 
    \centering
    \scalebox{0.9}{
    \begin{tabular}{lcc}
         & EM & F1\\
        \hline
        \hline
        Eqn.~\ref{eqn:sim}: dot product & 65.5 & 75.5\\
        Eqn.~\ref{eqn:sim}: linear & 59.5 & 69.7\\
        Eqn.~\ref{eqn:sim}: bilinear & 61.6 & 71.8 \\ 
        Eqn.~\ref{eqn:sim}: linear after MLP & 66.2 & 76.4 \\
        \hline
        Eqn.~\ref{eqn:flow}: MLP after concat & 67.1 & 77.0\\
        \hline
        \sysshort\ (single) & 68.0 & 77.3 \\
    \end{tabular}
    }
    \caption{Variations of similarity function $\alpha$ (Equation~\ref{eqn:sim}) and fusion function ${\bm \beta}$ (Equation~\ref{eqn:flow}) and their performance on the dev data of SQuAD. See Appendix~\ref{app:var} for the details of each variation.}
    \label{tab:var}
\end{table}

In this appendix section, we experimentally demonstrate how different choices of the similarity function $\alpha$ (Equation~\ref{eqn:sim}) and the fusion function ${\bm \beta}$ (Equation~\ref{eqn:flow}) impact the performance of our model.
Each variation is defined as following:

\paragraph{Eqn.~\ref{eqn:sim}: dot product.} 
Dot product $\alpha$ is defined as
\begin{equation}
\alpha({\bf h}, {\bf u}) = {\bf h}^\top {\bf u}
\end{equation}
where $\top$ indicates matrix transpose.
Dot product has been used for the measurement of similarity between two vectors by~\cite{hill2015goldilocks}.

\paragraph{Eqn.~\ref{eqn:sim}: linear.} 
Linear $\alpha$ is defined as
\begin{equation}
\alpha({\bf h}, {\bf u}) = {\bf w}_\textup{lin}^\top[{\bf h}; {\bf u}]
\end{equation}
where ${\bf w}^\top_\textup{lin} \in \mathbb{R}^{4d}$ is a trainable weight matrix.
This can be considered as the simplification of Equation~\ref{eqn:sim} by dropping the term ${\bf h} \circ {\bf u}$ in the concatenation. 

\paragraph{Eqn.~\ref{eqn:sim}: bilinear.} 
Bilinear $\alpha$ is defined as
\begin{equation}
\alpha({\bf h}, {\bf u}) = {\bf h}^\top {\bf W}_\textup{bi} {\bf u}
\end{equation}
where ${\bf W}_\textup{bi} \in \mathbb{R}^{2d \times 2d}$ is a trainable weight matrix. 
Bilinear term has been used by~\cite{thorough}.

\paragraph{Eqn.~\ref{eqn:sim}: linear after MLP.} 
We can also perform linear mapping after single layer of perceptron:
\begin{equation}
\alpha({\bf h}, {\bf u}) = {\bf w}_\textup{lin}^\top \tanh({\bf W}_\textup{mlp} [{\bf h}; {\bf u}] + {\bf b}_\textup{mlp})
\end{equation}
where ${\bf W}_\textup{mlp}$ and ${\bf b}_\textup{mlp}$ are trainable weight matrix and bias, respectively.
Linear mapping after perceptron layer has been used by~\cite{Hermann2015TeachingMT}.

\paragraph{Eqn.~\ref{eqn:flow}: MLP after concatenation.}
We can define ${\bm \beta}$ as
\begin{equation}
{\bm \beta}({\bf h}, \tilde{\bf u}, \tilde{\bf h}) = \max(0, {\bf W}_\textup{mlp} [{\bf h}; \tilde{\bf u}; {\bf h} \circ \tilde{\bf u}; {\bf h} \circ \tilde{\bf h}] + {\bf b}_\textup{mlp})
\end{equation}
where ${\bf W}_\textup{mlp} \in \mathbb{R}^{2d \times 8d}$ and ${\bf b}_\textup{mlp} \in \mathbb{R}^{2d}$ are trainable weight matrix and bias.
This is equivalent to adding ReLU after linearly transforming the original definition of ${\bm \beta}$.
Since the output dimension of ${\bm \beta}$ changes, the input dimension of the first LSTM of the modeling layer will change as well.

The results of these variations on the dev data of SQuAD are shown in Table~\ref{tab:var}.
It is important to note that there are non-trivial gaps between our definition of $\alpha$ and other definitions employed by previous work.
Adding MLP in ${\bm \beta}$ does not seem to help, yielding slightly worse result than ${\bm \beta}$ without MLP.


\end{document}